\documentclass[conference]{IEEEtran}
\IEEEoverridecommandlockouts
\usepackage{cite}
\usepackage{amsmath,amssymb,amsfonts}
\usepackage{algorithmic}
\usepackage{graphicx}
\usepackage{textcomp}
\usepackage{xcolor}
\usepackage{multirow}
\usepackage{subfigure}
\usepackage{pifont}
\usepackage[colorlinks,linkcolor=red]{hyperref}
\def\BibTeX{{\rm B\kern-.05em{\sc i\kern-.025em b}\kern-.08em
    T\kern-.1667em\lower.7ex\hbox{E}\kern-.125emX}}
\begin{document}

\title{Pyramid Real Image Denoising Network}



\author{\IEEEauthorblockN{Yiyun Zhao$^{1}$\qquad Zhuqing Jiang$^{1}$\qquad Aidong Men$^{1}$\qquad Guodong Ju$^{2}$}
\IEEEauthorblockA{$^{1}$\textit{Beijing University of Posts and Telecommunications}\qquad $^{2}$\textit{GuangDong TUS-TuWei Technology Co.,Ltd} \\
$^{1}$Beijing, China\qquad $^{2}$Guangzhou, China \\
$^{1}$\{yiyunzhao,\ jiangzhuqing,\ menad\}@bupt.edu.cn\qquad $^{2}$jgd@vip.163.com}}

\maketitle

\begin{abstract}
While deep Convolutional Neural Networks (CNNs) have shown extraordinary capability of modelling specific noise and denoising, they still perform poorly on real-world noisy images. The main reason is that the real-world noise is more sophisticated and diverse. To tackle the issue of blind denoising, in this paper, we propose a novel pyramid real image denoising network (PRIDNet), which contains three stages. First, the noise estimation stage uses channel attention mechanism to recalibrate the channel importance of input noise. Second, at the multi-scale denoising stage, pyramid pooling is utilized to extract multi-scale features. Third, the stage of feature fusion adopts a kernel selecting operation to adaptively fuse multi-scale features. Experiments on two datasets of real noisy photographs demonstrate that our approach can achieve competitive performance in comparison with state-of-the-art denoisers in terms of both quantitative measure and visual perception quality. Code is available at \url{https://github.com/491506870/PRIDNet}.
\end{abstract}


\section{Introduction}
Image denoising aims at restoring a clean image from its noisy one, which plays an essential role in low-level visual tasks. There has been considerable research on it, and near-optimal performances have been achieved for the removal of statistical distribution-regular noise (e.g., additive white Gaussian noise (AWGN), shot noise). Nevertheless, there is still a huge difference between specific noise and real-world noise. Among the latter, the noise comes from both shooting environment and image processing pipeline, 
thus its forms demonstrate complexity and diversity.

Recently deep Convolutional Neural Networks (CNNs) have led to significant improvements on denoising for specific noise. Mao et al. \cite{REDNet} present a very deep fully convolutional encoding-decoding framework with symmetric skip connections for Gaussian denoising, termed REDNet. Zhang et al. \cite{DnCNN} demonstrate that by merging residual learning and batch normalization, a denoising CNN (DnCNN) could outperform traditional non-CNN based methods. Other CNN methods \cite{N3Net,MemNet} also obtain promising denoising performance.

However, once the methods targeting for the specific noise above generalize to real-world noise, the performance may be even worse than the representative traditional methods such as BM3D \cite{BM3D}. Few blind denoising approaches especially for real noisy images are developed. By interactively setting relatively higher noise level, FFDNet \cite{FFDNet} can deal with more complex noise. CBDNet \cite{CBDNet} further utilizes a noise estimation subnetwork, so that the entire network could achieve end-to-end blind denoising.
\begin{figure}[tp]
\centering
\subfigure[Noisy Image]{
\includegraphics[width=0.22\linewidth]{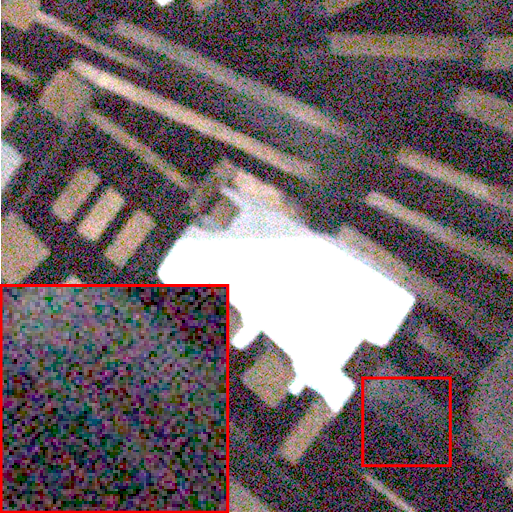}
}%
\subfigure[BM3D]{
\includegraphics[width=0.22\linewidth]{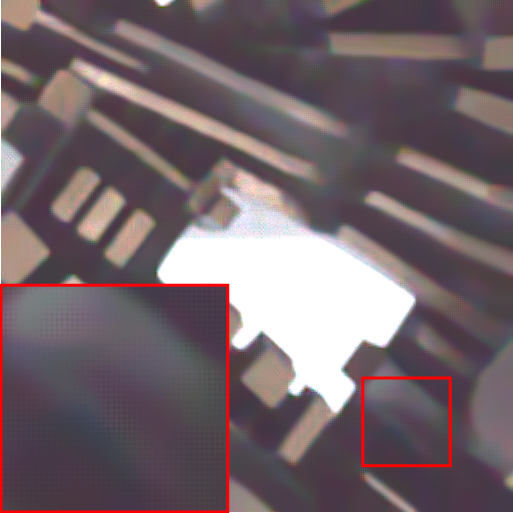}
}%
\subfigure[DnCNN]{
\includegraphics[width=0.22\linewidth]{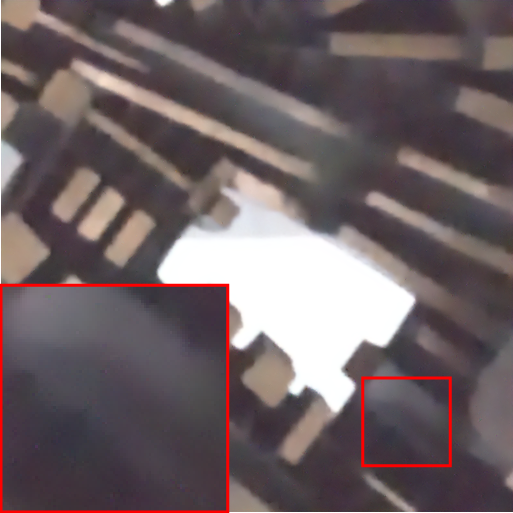}
}%
\subfigure[FFDNet]{
\includegraphics[width=0.22\linewidth]{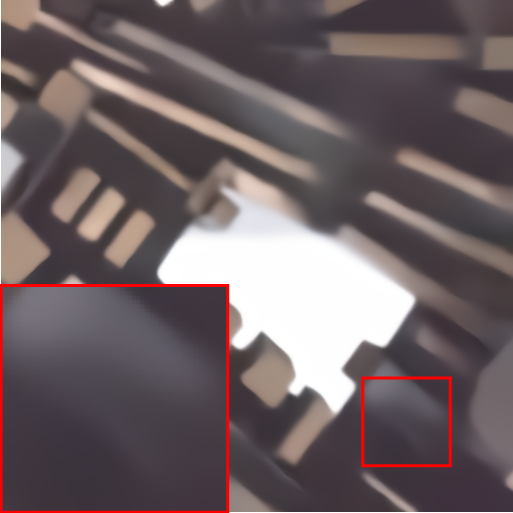}
}%
\quad
\subfigure[N3Net]{
\includegraphics[width=0.22\linewidth]{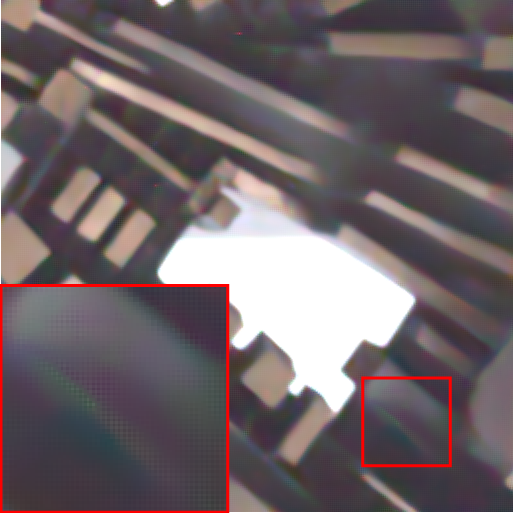}
}%
\subfigure[CBDNet]{
\includegraphics[width=0.22\linewidth]{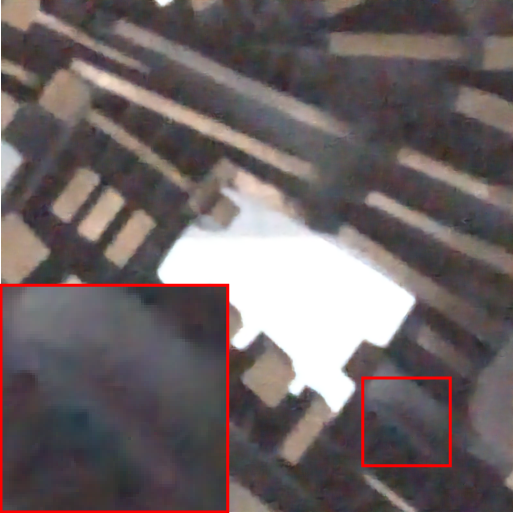}
}%
\subfigure[Path-Restore]{
\includegraphics[width=0.22\linewidth]{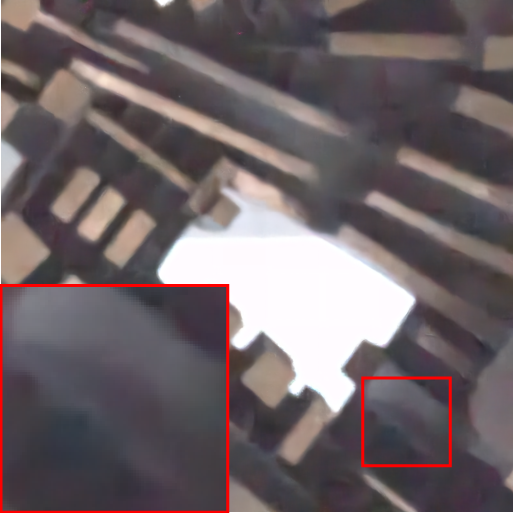}
}%
\subfigure[Ours]{
\includegraphics[width=0.22\linewidth]{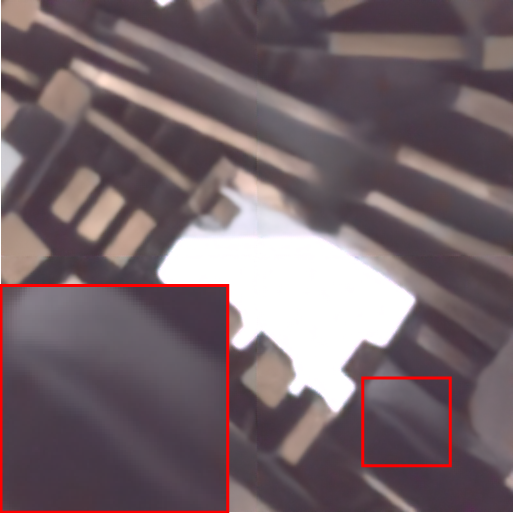}
}%

\centering
\caption{Denoising performance of state-of-the-art methods on a DND image. Readers are encouraged to zoom in for better visualization.}
\label{fig_DND}
\end{figure}
Yu et al. \cite{path-restore} propose a multi-path CNN named Path-Restore, which could dynamically select an appropriate route for each image region, especially for the varied noise distribution of a real noisy image.
As a commercial plug-in for Photoshop, Neat Image (NI) is also for bind denoising.

\begin{figure*}[ht]
\centering
\includegraphics[width=\textwidth]{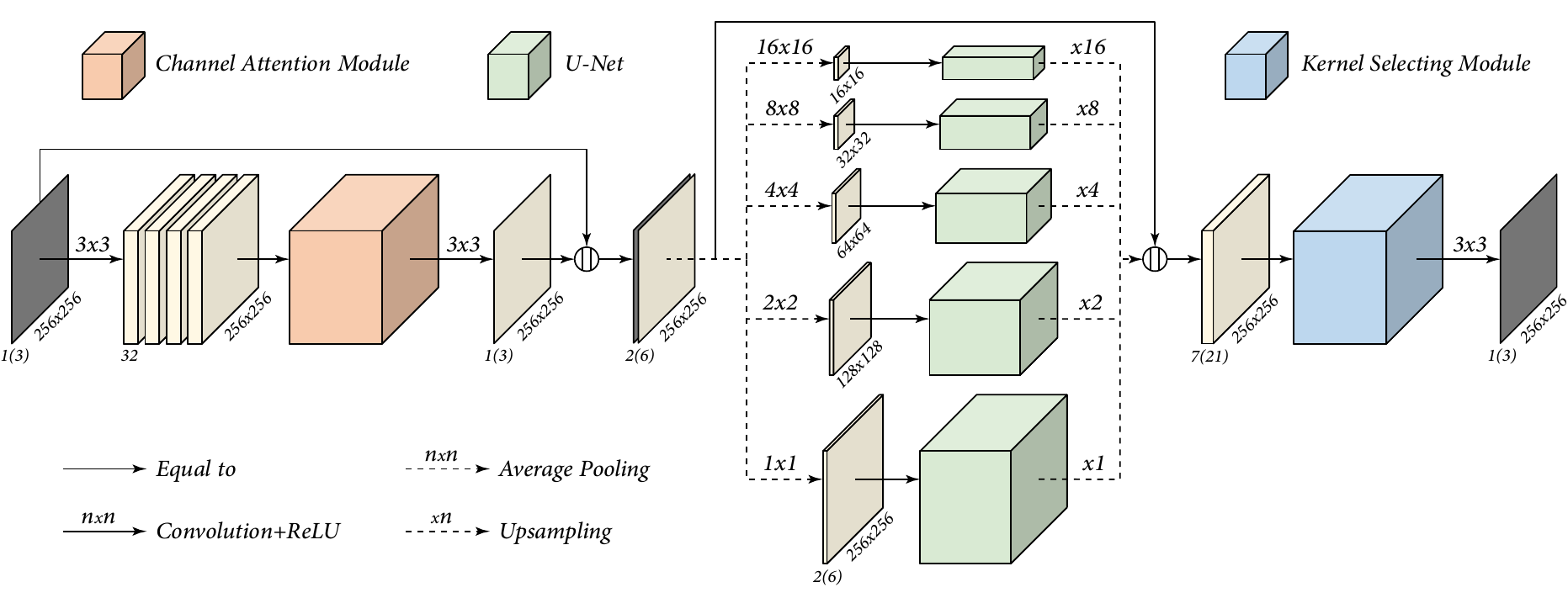}
\caption{The architecture of our proposed network (PRIDNet). The number of channels of feature maps is shown below them, for ``sRGB" model it is in the parentheses, while for ``raw" model it has no parentheses. The symbol $\parallel$ indicates concatenation.}
\label{fig_allnet}
\end{figure*}

Despite these methods have significantly improved the performance of real image denoising, there still remains three issues to be noticed:

First, in most of CNN based denoising methods, all channel-wise features are treated equally without adjustment according to their importance. In CNNs, different feature channels capture different types of noise across all regions of a single noisy image. Among them, some noise are more significant than others, thus should be assigned more weights.

Second, the previously mentioned methods, with fixed receptive fields, fail to carry diverse information. Referring to the traditional denoising method BM3D \cite{BM3D}, it searches for similar blocks in the whole image, taking the global information into consideration. Since features are not limited to a small area, receptive fields with different scales can fully exploit hierarchical spatial features. Context information would be very helpful when the image suffers from heavy noise.

Third, for the aggregation of multi-scale features, most of existing methods simply combine them in an element-wise summation manner or just concatenate them together \cite{REDNet}. Although containing the information of all scales, they treat features with different scales indiscriminately, ignoring the spatial and channel specificity of scale-wise features. Therefore multi-scale features can not be adaptively expressed.

To address these issues, we propose a pyramid real image denoising network (PRIDNet) as shown in Fig.~\ref{fig_allnet}. The main contribution of this work is three-fold:
\begin{itemize}
\item \emph{Channel attention:} Channel attention mechanism is utilized on the extracted noise features, adaptively recalibrating the channel importance.
\item \emph{Multi-scale feature extraction:} We design a pyramid denoising structure, in which each branch pays attention to one-scale features. Benefitted from it, we can extract global information and retain local details simultaneously, thereby making preparation for following comprehensive denoising.

\item \emph{Feature self-adaptive fusion:} Within concatenated multi-scale features, each channel represents one-scale features. We introduce a kernel selecting module. Multiple branches with different convolutional kernel sizes are fused by linear combination, allowing different feature maps to express by size-different kernels.
\end{itemize}

\section{proposed method}
In this section, we will formulate the proposed PRIDNet, including network architecture and three stages.
\subsection{Network Architecture}
As shown in Fig.~\ref{fig_allnet}, our model includes three stages: noise estimation stage, multi-scale denoising stage and feature fusion stage. The input noisy image is processed by three stages sequentially. Since all the operations are spatially invariant, it is robust enough to handle input images of arbitrary size.
To avoid loss of information, the output of the first stage is concatenated with its input before fed into next stage, likewise the second stage.
\subsection{Noise Estimation Stage}
This stage focuses on extracting discriminative features from input noisy images, which is regarded as an estimation of the noise level \cite{CBDNet}. We adopt a plain five-layer fully convolutional subnetwork without pooling and batch normalization, ReLU is deployed after each convolution. In each convolution layer, the number of feature channels is set to 32 (except the last layer is 1 or 3), and the filter size is 3$\times$3.

\begin{figure}
\centering{\includegraphics[width=\linewidth]{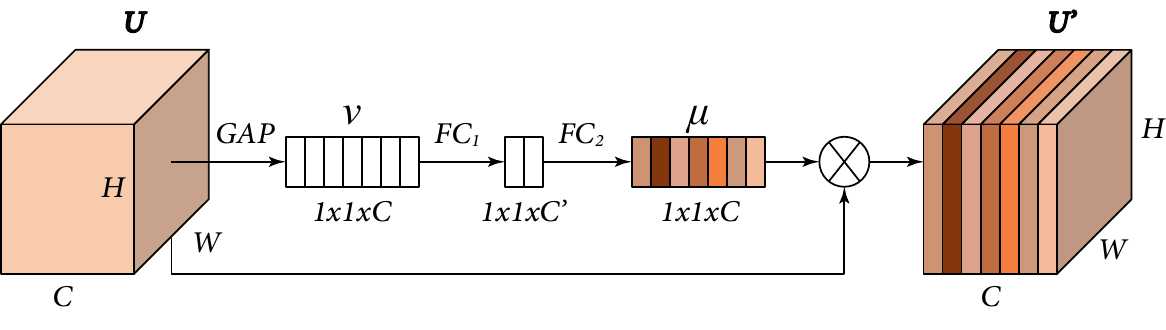}}
\caption{Channel attention module architecture. The \emph{GAP} denotes the global average pooling operation. \emph{$FC_1$} has a ReLU activation after it, and \emph{$FC_2$} has a Sigmoid activation after it. For concision, we omit these items.}
\label{fig_se}
\end{figure}

Before the last layer of stage, a channel attention module \cite{Luyue} is inserted to explicitly calibrate the interdependencies between feature channels. As shown in Fig.~\ref{fig_se}, the collection of channel weight $\mu=[\mu_1,\mu_2,...,\mu_c]\in\mathbb{R}^{1\times1\times{C}}$ is our goal, which is applied to rescale input feature maps $U\in\mathbb{R}^{H\times{W}\times{C}}$ to generate recalibrated features. We first squeeze global information of $U$ into a channel descriptor $\nu\in\mathbb{R}^{1\times1\times{C}}$ by using global average pooling (GAP). Then, it is followed by two fully connected layers (FC), and the number of channels in middle layer is set to 2. Above process can be formulated as
\begin{equation}
\mu=Sigmoid(FC_2(ReLU(FC_1(GAP(U))))).
\end{equation}
The final output of channel attention module (denoted as $U'\in\mathbb{R}^{H\times{W}\times{C}}$) is obtained by
\begin{equation}
U'=U\circ\mu,
\end{equation}
where $\circ$ refers to channel-wise multiplication between $U_{i}\in\mathbb{R}^{H\times{W}}$ and scalar calibration weight $\mu_{i}$, $i=1,2,...C$.
\subsection{Multi-scale Denoising Stage}
The idea of pyramid pooling is widely used in the fields of scene parsing \cite{pspnet}, image compression and so on. While to the best of our knowledge, it has never been used in the image denoising. Zhou et al. \cite{zhou} show that the empirical receptive field of CNN is much smaller than the theoretical one especially on high-level layers, which means that global information is not fully integrated when extracting features. On the contrary, for the removal of noise covering entire image, it has great help to match goal blocks with similar content in the whole image.

To mitigate this problem, we develop a five-level pyramid. Through five parallel ways, the input feature maps are downsampled to different sizes, helping branches gain relatively scale-different receptive fields to capture original, local and global information simultaneously. Pooling kernels are set to 1$\times$1, 2$\times$2, 4$\times$4, 8$\times$8 and 16$\times$16, respectively. Each pooled feature is then followed by U-Net \cite{u-net}, the network composed of deep encoding-decoding and skip connections, for studies have shown that successive upsampling and downsampling are helpful for denoising tasks. Note that five U-Nets do not share weights. At final of this stage, multi-level denoised features are upsampled by bilinear interpolation to the same size, and then concatenated together.
\subsection{Feature Fusion Stage}
\begin{figure}
\centering{\includegraphics[width=\linewidth]{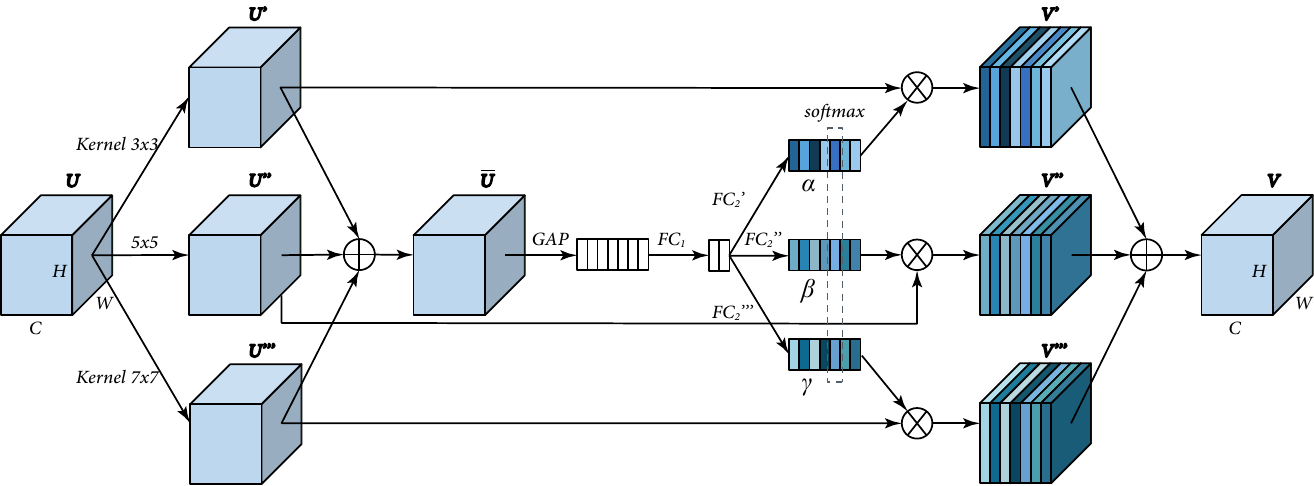}}
\caption{Kernel selecting module architecture. The \emph{GAP} denotes the global average pooling operation. A ReLU activation after \emph{$FC_1$} is omitted for brief.}
\label{fig_sk}
\end{figure}
In order to choose size-different kernel for each channel within concatenated multi-scale results, inspired by \cite{SKnet}, a kernel selecting module is introduced.
Details of a kernel selecting module is shown in Fig.~\ref{fig_sk}. The given feature maps $U\in\mathbb{R}^{H\times{W}\times{C}}$ are conducted by three parallel convolutions with kernel size 3, 5 and 7, respectively, to get $U'\in\mathbb{R}^{H\times{W}\times{C}}$, $U''\in\mathbb{R}^{H\times{W}\times{C}}$ and $U'''\in\mathbb{R}^{H\times{W}\times{C}}$. We first integrate information from all branches via element-wise summation:
\begin{equation}
\overline{U}=U'+U''+U'''.
\end{equation}
Then $\overline{U}$ is shrunk and then expanded by passing through a GAP and two FCs, the same operations as in the channel attention module, but no Sigmoid at last. The three outputs of $FC_2$, $\alpha'\in\mathbb{R}^{1\times1\times{C}}$, $\beta'\in\mathbb{R}^{1\times1\times{C}}$ and $\gamma'\in\mathbb{R}^{1\times1\times{C}}$ are operated by Softmax, which is applied across branches on the channel-wise, like gating mechanism:
\begin{equation}
k_c=\frac{e^{k'_c}}{e^{\alpha'_c}+e^{\beta'_c}+e^{\gamma'_c}},k=\alpha, \beta, \gamma,
\end{equation}
where $\alpha$, $\beta$ and $\gamma$ denote the soft attention vector for $U'$, $U''$ and $U'''$, respectively. Note that $\alpha_c$ is the c-th element of $\alpha$, likewise $\beta_c$ and $\gamma_c$. The final output feature maps $V$ are computed via combining various kernels with their attention weights:
\begin{equation}
V_c=\alpha_c\cdot{U'}+\beta_c\cdot{U''}+\gamma_c\cdot{U'''},
\end{equation}
where $\alpha$, $\beta$ and $\gamma$ need to satisfy $\alpha_c+\beta_c+\gamma_c=1$, and $V=[V_1,V_2,...,V_c]$, $V_c\in\mathbb{R}^{H\times{W}}$. Finally, we utilize a ${1\times1}$ convolutional layer to compress the dimension to 1 or 3 for feature fusion.
\section{experiments}
\subsection{Datasets}\label{AA}
For training, we utilize 320 image pairs (noisy and clean) both in raw-RGB space and sRGB space from Smartphone Image Denoising Dataset (SIDD) \cite{SIDD}. And we set another 1280 $256\times256$ crops of 40 images in SIDD as our validation data to conduct our ablation study.

For testing, we adopt two widely used benchmark datasets: DND \cite{DND} and NC12 \cite{NC}. DND, a benchmark of 50 real high-resolution images, captured by consumer grade cameras. Since only noisy images are provided to the public, PSNR/SSIM of denoised results are obtained through the online submission system\footnote{https://noise.visinf.tu-darmstadt.de/}. NC12 contains 12 noisy images, we only show the denoising results for qualitative evaluation as the ground-truth clean images are unavailable.
\subsection{Implementation Details}
We train two models, one targeting raw images and the other targeting sRGB images. The whole network is optimized with \emph{$L_1$} loss, and trained by Adam with $\beta_1=0.9$, $\beta_2=0.999$ and $\epsilon=10^{-8}$. All models are trained with 4000 epochs, where the learning rate for the first 1500 epochs is $10^{-4}$, and then $10^{-5}$ to finetune the model. We set patch size to $256\times256$, and batch size is set to 2 for ``raw'' model while 8 for ``sRGB'' model. All the experiments are implemented with TensorFlow on an NVIDIA GTX 1080ti GPU.
\subsection{Comparison with State-of-the-arts}
\begin{table}[tbp]
\caption{Quantitative performance of our model on DND compared with other published techniques, and sorted by sRGB PSNR.}
\begin{center}
\resizebox{\linewidth}{!}{
\begin{tabular}{c|cc|cc|c}
                       & \multicolumn{2}{c|}{Raw}         & \multicolumn{2}{c|}{sRGB}       &                   \\ \hline
Algorithm              & PSNR           & SSIM            & PSNR           & SSIM           & Blind / Non-blind \\ \hline\hline
TNRD \cite{TNRD}        & 44.97          & 0.9624          & 33.65          & 0.8306         & Non-blind         \\
BM3D \cite{BM3D}         & 46.64          & 0.9724          & 34.51          & 0.8507         & Non-blind         \\
KSVD \cite{KSVD}        & 45.54          & 0.9676          & 36.49          & 0.8978         & Non-blind         \\
WNNM \cite{WNNM}        & 46.30          & 0.9707          & 37.56          & 0.9313         & Non-blind         \\
FFDNet \cite{FFDNet}       & -              & -               & 37.61          & 0.9415         & Non-blind         \\
DnCNN \cite{DnCNN}        & -              & -               & 37.90          & 0.9430         & Blind             \\
CBDNet \cite{CBDNet}      & -              & -               & 38.06          & 0.9421         & Blind             \\
Path-Restore \cite{path-restore}& -              & -               & 39.00          & \textbf{0.9542}& Blind             \\
N3Net \cite{N3Net}        & 47.56          & 0.9767          & -              & -              & Blind             \\ \hline
PRIDNet                & \textbf{48.48} & \textbf{0.9806} & \textbf{39.42} & 0.9528         & Blind
\end{tabular}}
\label{table_DND}
\end{center}
\end{table}
\begin{table}
\caption{Ablation study of our model on validation dataset of SIDD.}
\begin{center}
\resizebox{\linewidth}{!}{
\begin{tabular}{ccc|c}
Channel Attention & Pyramid    & Kernel Selecting & SIDD  \\ \hline\hline
-                 & -          & -                & 51.81 \\
-                 & $\surd$    & $\surd$          & 52.02 \\
$\surd$           & -          & $\surd$          & 51.98 \\
$\surd$           & $\surd$    & -                & 52.15 \\
$\surd$           & $\surd$    & $\surd$          & \textbf{52.20}
\end{tabular}}
\label{table_ablation}
\end{center}
\end{table}
\textbf{Qualitative and Quantitative Results on DND Benchmark.}  Following the test protocol and tools provided by the website, we process 1000 bounding boxes in 50 real noisy images. The performance of our models with respect to prior work is shown in Table ~\ref{table_DND}. Note that we do not take unpublished or anonymous work into consideration although they have released results. We can see that although state-of-the-art non-blind Gaussian methods, e.g., TNRD \cite{TNRD}, BM3D \cite{BM3D} and WNNM \cite{WNNM}, are provided by noise level, they still have poorly performance, mainly due to the much difference between AWGN and real noise. CBDNet \cite{CBDNet} and Path-Restore \cite{path-restore} are especially trained for blind denoising of real images, thus yield promising results.
Our PRIDNet achieves large PSNR gains (i.e., $\sim0.9$dB for raw, and $\sim0.4$dB for sRGB) over the second best method. As for running time, PRIDNet takes about 0.05s to process an $512\times512$ image.

The visual denoising results by the various methods are shown in Fig.~\ref{fig_DND} and the supplementary material. CBDNet \cite{CBDNet} still contains some noise. DnCNN \cite{DnCNN} suffers from edge distortion. BM3D, N3Net and Path-Restore introduce some color artifacts to the denoised results, while FFDNet \cite{FFDNet} suffers from the problems of over-smoothing and loss of details and structures. Compared with these state-of-the-arts, our method (PRIDNet) achieves a better denoising performance by removing almost all the noise while preserving more fine textural details of whole image.

\textbf{Qualitative Comparisons on NC12.}  The methods we consider for comparison include blind denoising approaches (NC \cite{NC}, NI and CBDNet \cite{CBDNet}), blind Gaussian denoising approaches (DnCNN \cite{DnCNN}), and non-blind Gaussian denoising approaches (BM3D \cite{BM3D} and FFDNet \cite{FFDNet}). For non-blind methods, we exploit NI to estimate the noise level \emph{std.} so that they can perform the best visual quality. Due to limited space, we leave the visual comparisons of above methods in the supplementary material.

\subsection{Ablation Studies}
We conduct four ablation experiments to assess the importance of three key components in our PRIDNet. All ablation experiments are evaluated on the validation data of SIDD \cite{SIDD}. We can conclude from Table ~\ref{table_ablation} that the design of pyramid feature processing is the most crucial for our real image denoising task, which improves 0.22dB. The combination of three key components effectively boost network performance by 0.39dB compared with the plain network.
\section{Conclusion}
In this paper, a PRIDNet is presented for blind denoising of real images. The proposed network includes three sequential stages. The first stage explores relative importance of feature channels. At the second stage, pyramid pooling is developed to denoise multi-scale features. At the last stage, the operation of self-adaptive kernel selecting is introduced for feature fusion. Both qualitative and quantitative experiments show that our method achieves competitive performance.
\section*{Acknowledgments}
This work is supported by National Natural Science Foundation of China (No.61671077, 61671264) and the Fundamental Research Funds for the Central Universities 2019PTB-011.

\begin{figure*}[p]
\centering
\subfigure[Noisy Image]{
\includegraphics[width=0.22\textwidth]{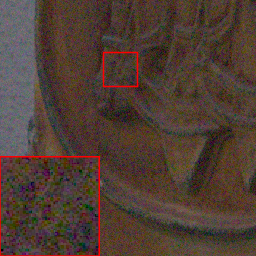}
}%
\subfigure[BM3D]{
\includegraphics[width=0.22\textwidth]{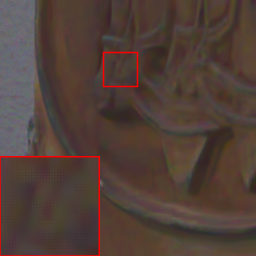}
}%
\subfigure[DnCNN]{
\includegraphics[width=0.22\textwidth]{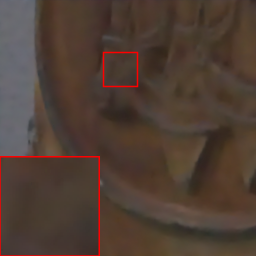}
}%
\subfigure[FFDNet]{
\includegraphics[width=0.22\textwidth]{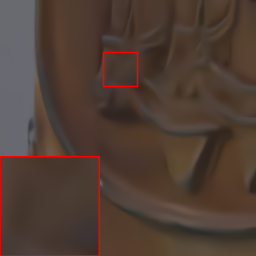}
}%
\quad
\subfigure[N3Net]{
\includegraphics[width=0.22\textwidth]{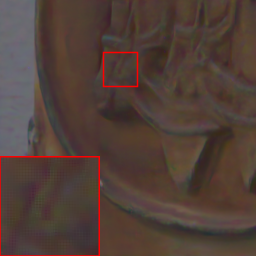}
}%
\subfigure[CBDNet]{
\includegraphics[width=0.22\textwidth]{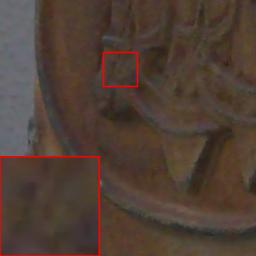}
}%
\subfigure[Path-Restore]{
\includegraphics[width=0.22\textwidth]{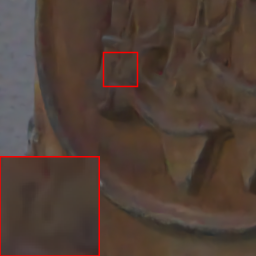}
}%
\subfigure[Ours]{
\includegraphics[width=0.22\textwidth]{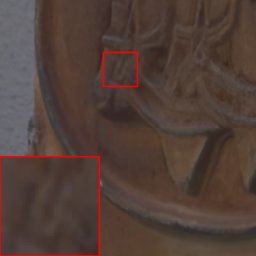}
}%

\centering
\caption{Denoising performance of state-of-the-art methods on another DND image. Readers are encouraged to zoom in for better visualization.}
\label{fig_DND-2}
\end{figure*}

\begin{figure*}[p]
\centering
\subfigure[Noisy Image]{
\includegraphics[width=0.22\textwidth]{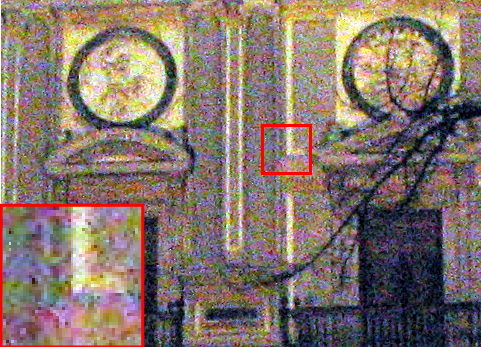}
}%
\subfigure[BM3D]{
\includegraphics[width=0.22\textwidth]{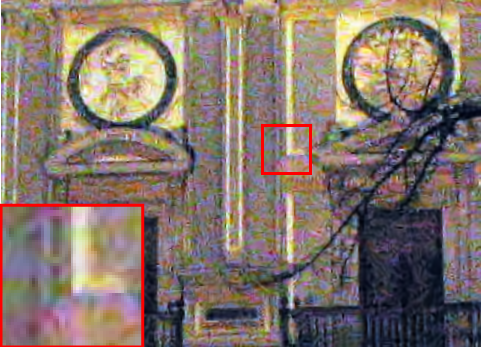}
}%
\subfigure[DnCNN]{
\includegraphics[width=0.22\textwidth]{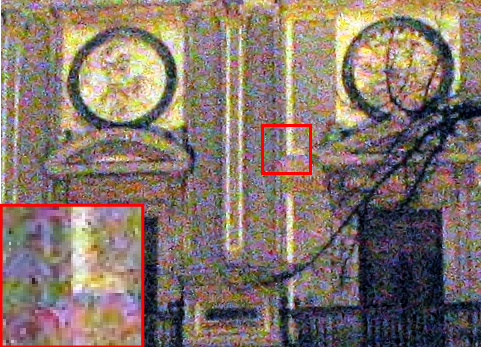}
}%
\subfigure[FFDNet]{
\includegraphics[width=0.22\textwidth]{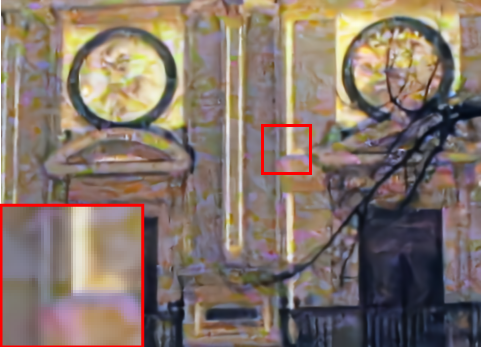}
}%
\quad
\subfigure[CBDNet]{
\includegraphics[width=0.22\textwidth]{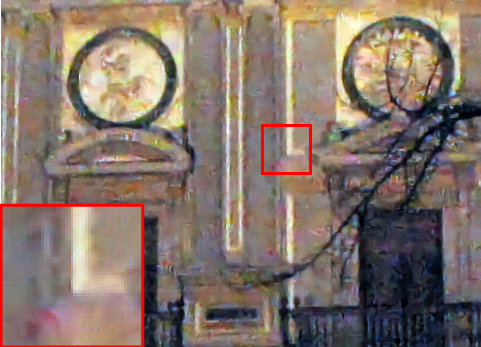}
}%
\subfigure[NC]{
\includegraphics[width=0.22\textwidth]{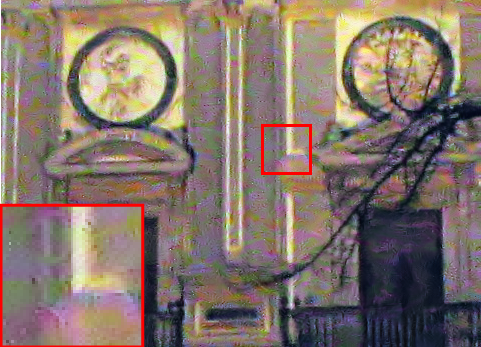}
}%
\subfigure[NI]{
\includegraphics[width=0.22\textwidth]{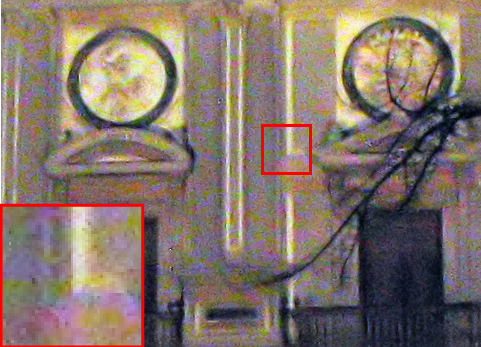}
}%
\subfigure[Ours]{
\includegraphics[width=0.22\textwidth]{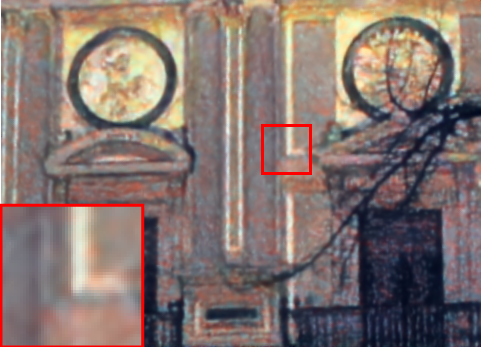}
}%

\centering
\caption{Denoising performance of state-of-the-art methods on a NC12 image. Readers are encouraged to zoom in for better visualization.}
\label{fig_NC12-1}
\end{figure*}

\bibliographystyle{IEEEbib}
\bibliography{VCIP}

\begin{thebibliography}{10}

\bibitem{REDNet}
X.-J Mao, C.-H Shen, and Y.-B Yang,
\newblock ``Image restoration using very deep convolutional encoder-decoder
  networks with symmetric skip connections,''
\newblock in {\em NIPS}, 2016, pp. 2802--2810.

\bibitem{DnCNN}
K.~Zhang, W.-M. Zuo, Y.-J. Chen, D.-Y. Meng, and L.~Zhang,
\newblock ``Beyond a gaussian denoiser: Residual learning of deep cnn for image
  denoising,''
\newblock {\em IEEE Transactions on Image Processing}, vol. 26, no. 7, pp.
  3142--3155, 2017.

\bibitem{N3Net}
T.~Pl{\"o}tz and S.~Roth,
\newblock ``Neural nearest neighbors networks,''
\newblock in {\em NIPS}, 2018, pp. 1087--1098.

\bibitem{MemNet}
Y.~Tai, J.~Yang, X.-M. Liu, and C.-Y. Xu,
\newblock ``Memnet: A persistent memory network for image restoration,''
\newblock in {\em ICCV}, 2017, pp. 4539--4547.

\bibitem{BM3D}
K.~Dabov, A.~Foi, V.~Katkovnik, and K.~O. Egiazarian,
\newblock ``Color image denoising via sparse 3d collaborative filtering with
  grouping constraint in luminance-chrominance space.,''
\newblock in {\em ICIP}, 2007, pp. 313--316.

\bibitem{FFDNet}
K.~Zhang, W.-M. Zuo, and L.~Zhang,
\newblock ``Ffdnet: Toward a fast and flexible solution for cnn-based image
  denoising,''
\newblock {\em IEEE Transactions on Image Processing}, vol. 27, no. 9, pp.
  4608--4622, 2018.

\bibitem{CBDNet}
S.~Guo, Z.-F. Yan, K.~Zhang, W.-M. Zuo, and L.~Zhang,
\newblock ``Toward convolutional blind denoising of real photographs,''
\newblock {\em arXiv preprint arXiv:1807.04686}, 2018.

\bibitem{path-restore}
K.~Yu, X.-T. Wang, C.~Dong, X.-O. Tang, and C.~C. Loy,
\newblock ``Path-restore: Learning network path selection for image
  restoration,''
\newblock {\em arXiv preprint arXiv:1904.10343}, 2019.

\bibitem{Luyue}
Y.~Lu, Z.-Q Jiang, G.-D Ju, L.-C Shen, and A.-D Men,
\newblock ``Recursive multi-stage upscaling network with discriminative fusion
  for super-resolution,''
\newblock in {\em ICME}, 2019, pp. 574--579.

\bibitem{pspnet}
H.-S Zhao, J.-P Shi, X.-J Qi, X.-G Wang, and J.-Y Jia,
\newblock ``Pyramid scene parsing network,''
\newblock in {\em CVPR}, 2017, pp. 2881--2890.

\bibitem{zhou}
B.-L. Zhou, A.~Khosla, A.~Lapedriza, A.~Oliva, and A.~Torralba,
\newblock ``Object detectors emerge in deep scene cnns,''
\newblock 2015.

\bibitem{u-net}
O.~Ronneberger, P.~Fischer, and T.~Brox,
\newblock ``U-net: Convolutional networks for biomedical image segmentation,''
\newblock in {\em MICCAI}, 2015, pp. 234--241.

\bibitem{SKnet}
X.~Li, W.-H. Wang, X.-L. Hu, and J.~Yang,
\newblock ``Selective kernel networks,''
\newblock {\em arXiv preprint arXiv:1903.06586}, 2019.

\bibitem{SIDD}
A.~Abdelhamed, S.~Lin, and M.~S. Brown,
\newblock ``A high-quality denoising dataset for smartphone cameras,''
\newblock in {\em CVPR}, 2018, pp. 1692--1700.

\bibitem{DND}
T.~Plotz and S.~Roth,
\newblock ``Benchmarking denoising algorithms with real photographs,''
\newblock in {\em CVPR}, 2017, pp. 1586--1595.

\bibitem{NC}
M.~Lebrun, M.~Colom, and J.~M. Morel,
\newblock ``The noise clinic: A blind image denoising algorithm,''
\newblock {\em IPOL}, vol. 5, pp. 1--54, 2015.

\bibitem{TNRD}
Y.-J. Chen, W.~Yu, and T.~Pock,
\newblock ``On learning optimized reaction diffusion processes for effective
  image restoration,''
\newblock in {\em CVPR}, 2015, pp. 5261--5269.

\bibitem{KSVD}
M.~Aharon, M.~Elad, A.~Bruckstein, et~al.,
\newblock ``K-svd: An algorithm for designing overcomplete dictionaries for
  sparse representation,''
\newblock {\em IEEE Transactions on signal processing}, vol. 54, no. 11, pp.
  4311, 2006.

\bibitem{WNNM}
S.-H. Gu, L.~Zhang, W.-M. Zuo, and X.-C. Feng,
\newblock ``Weighted nuclear norm minimization with application to image
  denoising,''
\newblock in {\em CVPR}, 2014, pp. 2862--2869.

\end{thebibliography}

\end{document}